\documentclass{article}

\usepackage{graphicx} 
\usepackage{subfigure} 

\usepackage{algorithm}
\usepackage{algorithmic}

\usepackage{times}
\usepackage{pdfsync}
\usepackage{comment}
\usepackage{amsmath}
\usepackage{amssymb}
\usepackage{amsthm}
\usepackage{calc}
\usepackage{color}
\usepackage{graphicx}
\usepackage{url}
\usepackage{xspace}
\usepackage{caption}
\usepackage{xspace}
\usepackage{color}
\usepackage{booktabs}

\usepackage[algo2e, noend, noline, linesnumbered]{algorithm2e}
\DontPrintSemicolon

\makeatletter
\newcommand{\pushline}{\Indp}
\newcommand{\popline}{\Indm}
\makeatother

\newcommand{\argmax}{\operatornamewithlimits{argmax}}

\newcommand{\cA}{\mathcal{A}}

\newcommand{\cG}{\mathcal{G}}

\newcommand{\cR}{\mathcal{R}}
\newcommand{\cS}{\mathcal{S}}
\newcommand{\cT}{\mathcal{T}}
\newcommand{\cZ}{\mathcal{Z}}
\newcommand{\hQ}{\hat{Q}}

\newcommand{\ie}{{\it i.e.,}~}

\newcommand{\MCTS}{{\sc MCTS}}
\newcommand{\Update}{{\sc Update}}
\newcommand{\Playout}{{\sc Playout}}
\newcommand{\Select}{{\sc Select}}
\newcommand{\Expand}{{\sc Expand}}
\newcommand{\Simulate}{{\sc Simulate}}

\definecolor{darkgreen}{RGB}{0,125,0}

\newcounter{mlNoteCounter}

\usepackage[margin=1.5in]{geometry}

\newcommand{\bea}{\begin{eqnarray}}
\newcommand{\eea}{\end{eqnarray}}

\newcommand{\bdf}{\begin{df}\em}
\newcommand{\edf}{\end{df}}

\newcommand{\ben}{\begin{enumerate}}
\newcommand{\een}{\end{enumerate}}

\definecolor{grey}{rgb}{0.4,0.4,0.4}
\definecolor{loselots}{rgb}{1,0.4,0.4}
\definecolor{losesome}{rgb}{1,0.6,0.6}
\definecolor{losebit}{rgb}{1,0.8,0.8}
\definecolor{tie}{rgb}{1,1,1}
\definecolor{winbit}{rgb}{0.8,1,0.8}
\definecolor{winsome}{rgb}{0.6,1,0.6}
\definecolor{winlots}{rgb}{0.4,1,0.4}

\title{Monte Carlo Tree Search with Heuristic Evaluations\\using Implicit Minimax Backups}

\author{Marc Lanctot$^1$, Mark H.M. Winands$^1$, Tom Pepels$^1$, Nathan R. Sturtevant$^2$ \\
\hspace{-0.3cm}$^1$Games and AI Group, Department of Knowledge Engineering, Maastricht University\\
\hspace{-0.5cm}$^2$Computer Science Department, University of Denver\\
\vspace{0.3cm}
\hspace{-0.5cm}{\small\{marc.lanctot,m.winands,tom.pepels\}@maastrichtuniversity.nl, sturtevant@cs.du.edu}}

\date{}

\newif\iftechreport
\techreporttrue

\begin{document} 


\maketitle
\allowdisplaybreaks

\begin{abstract} 
Monte Carlo Tree Search (MCTS) has improved the performance of game engines in 
domains such as Go, Hex, and general game playing. MCTS has been shown to outperform
classic $\alpha\beta$ search in games where good heuristic evaluations are difficult to obtain. 
In recent years, combining ideas from traditional minimax search in MCTS has been shown to be advantageous in some domains, 
such as Lines of Action, Amazons, and Breakthrough.
In this paper, we propose 
a new way to use heuristic evaluations to guide the MCTS search by storing the two sources of 
information, estimated win rates and heuristic evaluations, separately. 
Rather than using the heuristic evaluations to replace the playouts, 
our technique backs them up {\it implicitly} during the MCTS simulations. 
These minimax values are then used to guide future simulations. 
We show that using implicit minimax backups  
leads to stronger play performance in Kalah, Breakthrough, and Lines of Action. 
\end{abstract} 

%
%

\section{Introduction}

Monte Carlo Tree Search (MCTS)~\cite{Coulom06Efficient,Kocsis06Bandit} is a simulation-based best-first
search technique that has been shown to increase performance in domains such as turn-taking games, 
general-game playing, real-time strategy games, single-agent planning, and more~\cite{mctssurvey}. 
While the initial applications have been to games where heuristic evaluations are difficult to obtain, 
progress in MCTS research has shown that heuristics can be effectively be combined in MCTS, even in games 
where classic minimax search has traditionally been preferred. 

The most popular MCTS algorithm is UCT~\cite{Kocsis06Bandit}, 
which performs a single simulation from the root of the search tree to a terminal state at each iteration. 
During the iterative process, a game tree is incrementally built by adding a 
new leaf node to the tree on each iteration, whose nodes maintain statistical estimates such as average payoffs. 
With each new simulation, these estimates improve and help to guide future simulations. 


In this work, we propose a new technique to augment the quality of MCTS simulations with  
an implicitly-computed minimax search which uses heuristic evaluations. 
Unlike previous work, these heuristic evaluations are used as {\it separate source of information}, 
and backed up in the same way as in classic minimax search. Furthermore, these minimax-style 
backups are done {\it implicitly},
as a simple extra step during the standard updates to the tree nodes, and always maintained 
separately from win rate estimates obtained from playouts. These two separate information 
sources are then used to guide MCTS simulations. 
We show that combining heuristic evaluations in this way can lead to significantly stronger play performance in three 
separate domains: Kalah, Breakthrough, and Lines of Action. 

\subsection{Related Work}

Several techniques for minimax-influenced backup rules in the simulation-based MCTS framework have been previously proposed. 
The first was Coulom's original {\it maximum backpropagation}~\cite{Coulom06Efficient}. This method of backpropagation
suggests, after a number of simulations to a node has been reached, to switch to propagating the maximum value instead 
of the simulated (average) value. 
The rationale behind this choice is that after a certain point, the search algorithm should consider a node
{\it converged} and return an estimate of the best value. 
Maximum backpropagation has also recently been used in other Monte Carlo search algorithms and demonstrated success in
probabilistic planning, as an alternative type of forecaster in BRUE~\cite{Feldman13Theoretically} and as Bellman 
backups for online dynamic programming in Trial-based Heuristic Tree Search~\cite{Keller13Trial}.

The first use of enhancing MCTS using prior knowledge was in Computer Go~\cite{Gelly07Combining}. 
In this work, offline-learned knowledge initialized values of expanded nodes increased performance against a significantly 
strong benchmark player. 
This technique was also confirmed to be advantageous in Breakthrough~\cite{Lorentz13Breakthrough}. 
Another way to introduce prior knowledge is via a progressive bias during selection~\cite{Chaslot08Progressive}, which has 
significantly increased performance in Go play strength~\cite{Chaslot10Adding}. 

In games where minimax search performs well, such as Kalah, 
modifying MCTS to use minimax-style backups and heuristic values instead to replace playouts offers a worthwhile trade-off 
under different search time settings~\cite{Ramanujan11Tradeoffs}.
Similarly, there is further evidence suggesting not replacing the playout entirely, but terminating them early 
using heuristic evaluations, has increased the performance in Lines of Action (LOA)~\cite{Winands10MCTS-LOA}, 
Amazons~\cite{Kloetzer10Amazons,Lorentz08Amazons}, and Breakthrough~\cite{Lorentz13Breakthrough}. In LOA and Amazons, the 
MCTS players enhanced with evaluation functions outperform their minimax counterparts using the same evaluation function.


One may want to combine minimax backups or searches without using an evaluation function. 
The prime example is MCTS-Solver~\cite{Winands08Solver}, which backpropagates proven wins and losses as 
extra information in MCTS. When a node is proven to be a 
win or a loss, it no longer needs to be searched. This domain-independent modification greatly enhances 
MCTS with negligible overhead. Score-bounded MCTS extends this idea to games with multiple 
outcomes, leading to $\alpha \beta$-style pruning in the tree~\cite{Cazenave10Score}. One can use shallow-depth
minimax searches in the tree to initialize nodes during expansion, enhance the playout, or to help MCTS-Solver 
in backpropagation~\cite{Baier13MinimaxHybrids}.

Finally, recent work has attempted to explain and identify some of the shortcomings that arise from estimates in 
MCTS, specifically compared to situations where classic minimax search has historically performed 
well~\cite{Ramanujan10Understanding,Ramanujan10On}. 
Attempts have been made to overcome the problem of {\it traps} or {\it optimistic moves}, \ie moves that initially seem 
promising but then later prove to be bad, such as sufficiency 
thresholds~\cite{Gudmundsson13Sufficiency} and shallow minimax searches~\cite{Baier13MinimaxHybrids}. 


\section{Adversarial Search in Turn-Taking Games}

A finite deterministic Markov Decision Process (MDP) is 4-tuple $(\cS, \cA, \cT, \cR)$. Here, $\cS$ is a finite non-empty set of {\it states}. 
$\cA$ is a finite non-empty set of {\it actions}, where we denote $\cA(s) \subseteq \cA$ the set of available actions at state $s$. 
$\cT : \cS \times \cA \mapsto \Delta \cS$ is a {\it transition function} mapping 
each state and action to a distribution over successor states. Finally, $\cR : \cS \times \cA \times \cS \mapsto R$ 
is a {\it reward function} mapping (state, action, successor state) triplets to numerical rewards. 

A two-player perfect information game is an MDP with a specific form.
Denote $\cZ = \{ s \in \cS: \cA(s) = \emptyset \} \subset \cS$ the set of {\it terminal states}. 
In addition, for all nonterminal states $s' \in \cS - \cZ$, $\cR(s,a,s') = 0$. 
There is a {\it player identity function} $\tau : \cS - \cZ \mapsto \{1,2\}$. 
The rewards $\cR(s,a,s')$ are always with respect to the same player and  
we assume zero-sum games so that rewards with respect to the opponent player are simply negated. 
In this paper, we assume fully deterministic domains, so $\cT(s,a)$ maps $s$ to a single successor 
state. 
However, the ideas proposed can be easily extended to domains with stochastic transitions. 
When it is clear from the context and unless otherwise stated, we denote $s' = \cT(s,a)$. 

Monte Carlo Tree Search (MCTS) is a simulation-based best-first search algorithm that incrementally builds a tree, $\cG$, 
in memory. 
Each search starts with from a {\it root state} $s_0 \in \cS - \cZ$, and initially sets $\cG = \emptyset$. 
Each simulation samples a trajectory $\rho = (s_0, a_0, s_1, a_1, \cdots, s_n)$, where $s_n \in \cZ$ unless the playout 
is terminated early. 
The portion of the $\rho$ where $s_i \in \cG$ is called the {\it tree portion} and the remaining portion is
called the {\it playout portion}. In the tree portion, actions are chosen according to some {\it selection policy}. 
The first state encountered in the playout portion is {\it expanded}, added to $\cG$.
The actions chosen in the playout portion are determined by a specific {\it playout policy}. 
States $s \in \cG$ are referred to as {\it nodes} and statistics are  
maintained for each node $s$: the cumulative reward, $r_s$, and visit count, $n_s$. 
By popular convention, we define $r_{s,a} = r_{s'}$ where $s' = \cT(s,a)$, and similarly $n_{s,a} = n_{s'}$. 
Also, we use $r^{\tau}_s$ to denote the reward at state $s$ {\it with respect to player} $\tau(s)$. 

Let $\hQ(s,a)$ be an estimator for the value of state-action pair $(s,a)$, where $s \in \cA(s)$. 
One popular estimator is the observed mean 
$Q(s,a) = r^{\tau}_{s,a} / n_{s,a}$. 
The most widely-used selection policy is based on a bandit algorithm called Upper Confidence Bounds 
(UCB)~\cite{Auer02Finite}, used in adaptive multistage sampling~\cite{Chang2005AMS} and in 
UCT~\cite{Kocsis06Bandit}, which selects action $a'$ using
\begin{equation}
\label{eq:select-ucb}
a' = \argmax_{a \in \cA(s)} \left\{ \hQ(s,a) + C \sqrt{\frac{\ln n_s}{n_{s,a}}} \right\}, 
\end{equation}
where $C$ is parameter determining the weight of exploration. 

\section{Implicit Minimax Backups in MCTS}

Our proposed technique is based on the following principle: if an evaluation function is available, then it should 
be possible to augment MCTS to make use of it for a potential gain in performance.  
Suppose we are given an evaluation function $v_0(s)$ whose range is the same as that of the reward function $\cR$. 
How should MCTS make use of this information? 
We propose a simple and elegant solution: add another value to maintain at each node, the 
{\it implicit minimax evaluation with respect to player} $\tau(s)$, $v^{\tau}_s$, with $v^{\tau}_{s,a}$ defined similarly 
as above. 
This new value at node $s$ {\it only} maintains a heuristic minimax value built from the evaluations of subtrees below $s$. 
During backpropagation, $r_s$ and $n_s$ are updated in the usual way, and additionally $v^{\tau}_s$ is updated using minimax backup 
rule based on children values. Then, similarly to RAVE~\cite{Gelly07Combining}, rather than using $\hQ = Q$ for 
selection in Equation~\ref{eq:select-ucb}, we use
\begin{equation}
\label{eq:imq}
\hQ^{\mathit{IM}}(s,a) = (1-\alpha) \frac{r^{\tau}_{s,a}}{n_{s,a}} + \alpha v^{\tau}_{s,a}, 
\end{equation}
where $\alpha$ weights the influence of the heuristic minimax value.

The entire process is summarized in Algorithm~\ref{alg}. There are a few simple additions to standard MCTS,
located on lines 2, 8, 13, and 14.
During selection, $\hQ^{\mathit{IM}}$ from Equation~\ref{eq:imq} replaces $Q$ in 
Equation~\ref{eq:select-ucb}. During backpropagation, the implicit minimax evaluations $v^{\tau}_s$ are updated based on 
the children's values. For simplicity, a single $\max$ operator is used here since the evaluations are assumed to be in 
view of player $\tau(s)$. 
Depending on the implementation, the signs of rewards may depend on $\tau(s)$ and/or $\tau(s')$.
For example, a negamax implementation would include sign inversions at the appropriate places 
to ensure that the payoffs are in view of the current player at each node.  
Finally, \Expand\xspace adds all children nodes to the tree, sets their implicit minimax values to their initial
heuristic values on line 13, and does a one-ply backup on line 14. 
A more memory-efficient implementation could add just a single child without fundamentally changing
the algorithm, as was done in our experiments in Lines of Action.

\begin{algorithm2e}[t!]
  \Select$(s)$:\;
  \pushline
    Let $A'$ be the set of actions $a \in \cA(s)$ maximizing $\hQ^{\mathit{IM}}(s,a) + C \sqrt{\frac{\ln n_s}{n_{s,a}}}$ \; 
    {\bf return} $a' \sim ${\sc Uniform}$(A')$ \;
  \popline
  \;
  \Update$(s,r)$:\;
  \pushline
    $r_s \leftarrow r_s + r$\;
    $n_s \leftarrow n_s + 1$\;
    $v^{\tau}_s \leftarrow \max_{a \in \cA(s)} v^{\tau}_{s,a}$ \; 
  \popline
  \;
  \Simulate$(s_{parent}, a_{parent}, s)$:\;
  \pushline
    \If{$\exists a \in \cA(s), s' = \cT(s,a) \not\in \cG$}{
      \Expand($s$)\;      
      \lFor{$a \in \cA(s), s' = \cT(s,a)$}{$v_{s'} \leftarrow v_0(s')$ \;}  
      $v^{\tau}_s \leftarrow \max_{a \in \cA(s)} v^{\tau}_{s,a}$ \; 
      $r \leftarrow $\Playout($s$)\;
      \Update($s,r$)\;
      {\bf return} $r$\;                       
    }
    \Else{ 
      \lIf{$s \in \cZ$}{{\bf return} $\cR(s_{parent}, a_{parent}, s)$ \;}  
      $a \leftarrow $\Select$(s)$\;
      $s' \leftarrow \cT(s,a)$\;
      $r \leftarrow $\Simulate$(s,a,s')$\;
      \Update($s,r$)\;
      {\bf return} $r$\;                      
    }
  \popline
  \;
  \MCTS($s_0$):\;
  \pushline
    \lWhile{\mbox{time left}}{\Simulate$(-,-,s_0)$ \;}           
    {\bf return} $\argmax_{a \in \cA(s_0)} n_{s_0,a}$\; 
  \popline
  \vspace{0.3cm}
  \caption{MCTS with implicit minimax backups. \label{alg}}
\end{algorithm2e}

In essence, this defines a new {\it information scheme} where each node is augmented with 
heuristic estimates which are backed-up differently than the Monte Carlo statistics.
When MCTS-Solver is enabled, proven values take precedence in the selection policy 
and the resulting scheme is informative and consistent~\cite{Saffidine13PhdThesis}, so 
Algorithm \ref{alg} converges to the optimal choice eventually. However, before a node 
becomes a proven win or loss, 
the implicit minimax values act like an heuristic approximation of MCTS-Solver for the portion of the
search tree that has not reached terminal states. 



\section{Empirical Evaluation}

\newcommand{\UCTMAXH}{$\mbox{UCTMAX}_H$\xspace}
\newcommand{\UCTH}{$\mbox{UCT}_H$\xspace}

In this section, we thoroughly evaluate the practical performance of the implicit minimax backups technique. 
Before reporting head-to-head results, we first describe our experimental setup and 
summarize the techniques that have been used to improve playouts. We then present results on three game
domains: Kalah, Breakthrough, and Lines of Action. 

Unless otherwise stated, our implementations expand a new node every simulation, the first node encountered
that is not in the tree. MCTS-Solver is enabled in all of the experiments since its overhead is negligible and
never decreases performance. After the simulations, the move with the highest visit count 
is chosen on line 28. 
Rewards are in $\{-1, 0, 1\}$ representing a loss, draw, and win.
Evaluation function values are scaled to $[-1,1]$ by passing a domain-dependent 
score differences through a cache-optimized sigmoid function. 
When simulating, a single game state is modified and 
moves are undone when returning from the recursive call.
Whenever possible, evaluation functions are updated incrementally. 
All of the experiments include swapped seats to ensure that each player type plays 
an equal number of games as first player and as second player.
All reported win rates are over 1000 played games and search time is set to 1 second unless specifically stated 
otherwise.
Domain-dependent playout policies and optimizations are reported in each subsection.

We compare to and combine our technique with a number of other ones to include  
domain knowledge. A popular recent technique is {\it early playout terminations}. When a leaf node of the tree 
is reached, a fixed-depth early playout termination, hereby abbreviated to ``fet$x$'', plays $x$ moves according
to the playout policy resulting in state $s$, and then terminates the playout returning $v_0(s)$. This method has
shown to improve performance against standard MCTS in Amazons, Kalah, and 
Breakthrough~\cite{Lorentz13Breakthrough,Ramanujan11Tradeoffs,Lorentz08Amazons}. 

A similar technique is {\it dynamic early terminations}, which periodically checks the evaluation function 
(or other domain-dependent features) terminating only when some condition is met. 
This approach has been used as a ``mercy rule'' in Go~\cite{Bouzy07Old} and quite successfully in 
Lines of Action~\cite{Winands08MCTSSolver}.
In our version, which we abbreviate ``det$x$'', a playout is terminated and returns $1$ if $v_0(s) \ge x$ and 
$-1$ if $v_0(s) \le -x$. Another option is to use an $\epsilon$-greedy playout policy that chooses a successor randomly 
with probability $\epsilon$ and successor state with the largest evaluation with probability $1-\epsilon$, with 
improved performance in Chinese Checkers~\cite{Sturtevant08An,Nijssen12Playout}, abbreviated ``ege$\epsilon$''.

To facilitate the discussion, 
we refer to each enhancement and setting using different labels. These enhancements and labels are described in 
the text that follows. But, we also include, for reference, a summary of each in Table~\ref{table:enhancements}.

Experiments are performed in three domains: Kalah, Breakthrough, and Lines of Action. Example images of each 
game are shown in Appendix A.
To tune parameters in Kalah and Breakthrough, hierarchical elimination tournaments are run where each 
head-to-head match consisted of at least 200 games with seats swapped halfway. 
Detailed results of these tournaments and comparisons are contained in Appendix B.

\begin{table}[tb]
{\small
\caption{Enhancements tested in Kalah (K), Breakthrough (B), and Lines of Action (L).}
\begin{center}
\begin{tabular}{|l|c|c|c|c|}
\hline 
Enhancement / Setting       & Abbr.          & K           & B           & L \\ 
\hline                                                          
Improved playout policy     & ipp            &             & \checkmark  & \checkmark \\ 
Early playout termination   & fet$x$         & \checkmark  & \checkmark  &            \\
Dynamic early termination   & det$x$         &             & \checkmark  & \checkmark \\
$\epsilon$-greedy playouts  & ege$\epsilon$  &             & \checkmark  &            \\
Node priors                 & np             &             & \checkmark  &            \\
Maximum backpropagation     &                &             & \checkmark  &            \\
Progressive bias            & PB             &             & \checkmark  & \checkmark \\
$\alpha\beta$ playouts      &                &             &             & \checkmark \\
\hline                                                                   
Implicit minimax backups    & im$\alpha$     & \checkmark  & \checkmark  & \checkmark \\
\hline                                                                   
Simple evaluation function  & efRS, efMS     & \checkmark  & \checkmark  &            \\
Sophisticated ev. function  & efLH, efWB     &             & \checkmark  & \checkmark \\
Baseline pl. (ege$0.1$,det$0.5$)  & bl   &             & \checkmark  &            \\
Alt. baseline (ipp,fet$20$,np)           & bl'  &             & \checkmark  &            \\
\hline
\end{tabular}
\end{center} 
\label{table:enhancements} }
\end{table}%

\subsection{Kalah}

\begin{figure}
\begin{center}
\includegraphics[scale=0.68]{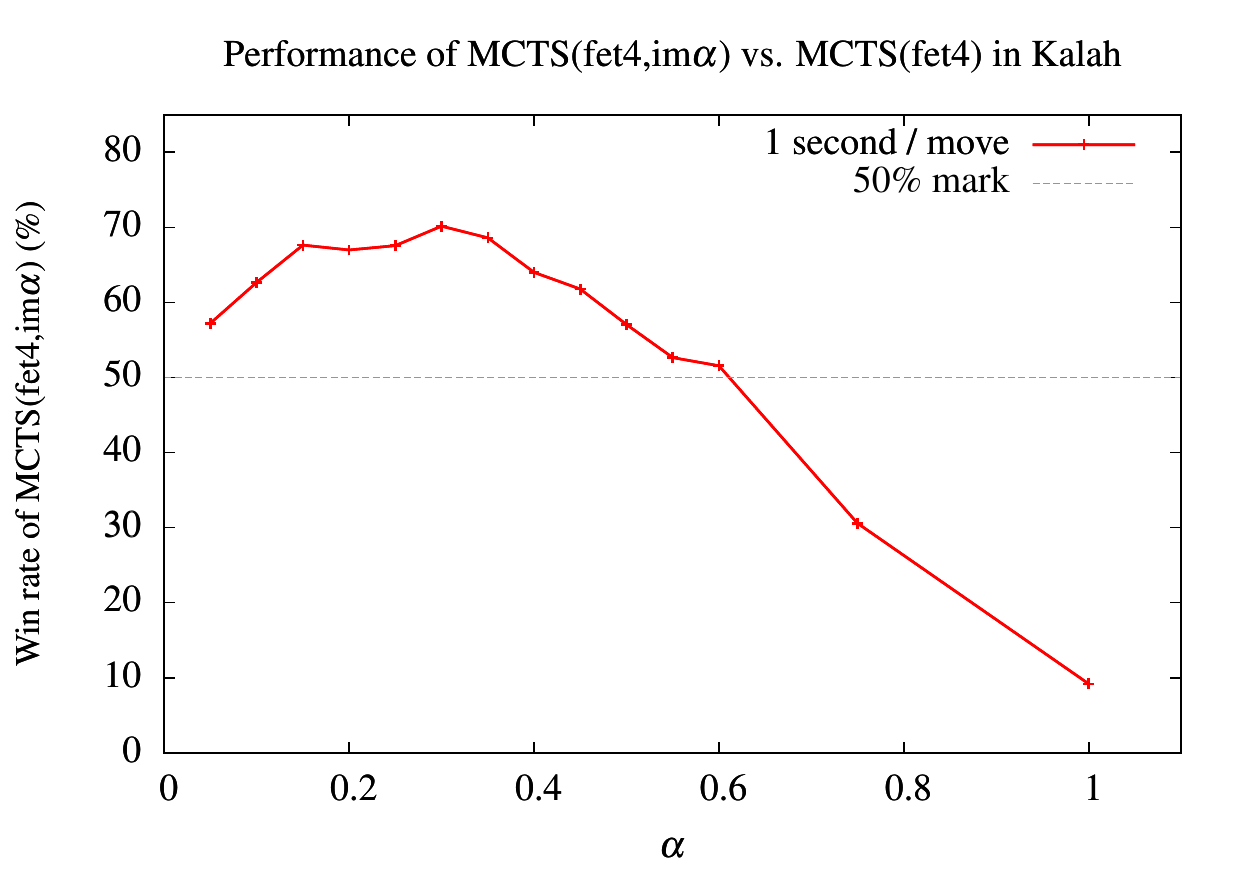}
\caption{Results in Kalah. Playouts use fet$4$. Each data point is based on roughly 1000 games.}
\label{fig:kalah-alpha}
\end{center}
\end{figure}

Kalah is a turn-taking game in the Mancala family of games. Each player has six houses, each 
initially containing four stones, and a store on the endpoint of the board, initially empty. 
On their turn, a player chooses one of their houses, removes all the stones in it, and ``sows'' 
the stones one per house in counter-clockwise fashion, skipping the opponent's store. 
If the final stone lands in the player's store, that player gets another turn, and there is no 
limit to the number of consecutive turns taken by same player. If the stone ends on a house owned
by the player that contains no stones, then that player captures all the stones in the adjacent 
opponent house, putting it into the player's store. The game plays until one player's houses are
all empty; the opponent then moves their remaining stones to their store. The winner is the player
who has collected the most stones in their store. 
Kalah has been weakly solved for several different variants of Kalah~\cite{Irving00Solving}, 
and was used as a domain to compare MCTS variants to classic minimax search~\cite{Ramanujan11Tradeoffs}.

In running experiments from the initial position, we observed a noticeable first-player bias. Therefore, as
was done in \cite{Ramanujan11Tradeoffs}, our experiments produce random starting board positions 
without any stones placed in the stores. 
Competing players play one game and then swap seats to play a second game using the same board. A player 
is declared a winner if that player won one of the games and at least tied the other game. If the same side 
wins both games, the game is discarded. 

The default playout policy chooses a move uniformly at random. 
We determined which playout enhancement led to the best player. 
Tournament results revealed that a fet$4$ early termination worked best. 
The evaluation function was the same one used in \cite{Ramanujan11Tradeoffs}, the difference between stones
in each player's stores. Results with one 
second of search time are shown in Figure~\ref{fig:kalah-alpha}. 
Here, we notice that within the range $\alpha \in [0.1,0.5]$ there is a clear 
advantage in performance when using implicit minimax backups against the base player. 

\subsection{Breakthrough}
\label{sec:bt}

Breakthrough is a turn-taking alternating move game played on an 8-by-8 chess board. Each player 
has 16 identical pieces on their first two rows. 
A piece is allowed to move forward to an empty square, either straight or diagonal, but may only 
capture diagonally like Chess pawns. A player wins by moving a single piece to the furthest opponent row. 

Breakthrough was first introduced in general game-playing competitions and has been identified as a domain 
that is particularly difficult for MCTS due to traps and uninformed playouts~\cite{Gudmundsson13Sufficiency}. 
Our playout policy always chooses one-ply ``decisive'' wins and prevents immediate ``anti-decisive'' 
losses~\cite{Teytaud10On}.
Otherwise, a move is selected non-uniformly at random, where capturing undefended pieces are four times more
likely than other moves. 
MCTS with this {\it improved playout policy} (abbreviated ``ipp'') beats the one using uniform random 
94.3\% of the time. This playout policy leads to a clear improvement over random playouts, and so it is enabled 
by default from this point on.

In Breakthrough, two different evaluation functions were used. The first one is a simple one found in 
Maarten Schadd's thesis~\cite{Schadd11PhdThesis} 
that assigns each piece a score of 10 and the further row achieved as 2.5, which we abbreviate ``efMS''. The second 
one is the more sophisticated one giving specific point values for each individual square per player 
described in a recent paper by Lorentz \& Horey~\cite{Lorentz13Breakthrough}, which we abbreviate ``efLH''. 
We base much of our analysis in Breakthrough on the Lorentz \& Horey player, which 
at the time of publication had an ELO rating of 1910 on the Little Golem web site. 


\begin{figure}[t]
\begin{center}
\includegraphics[scale=0.7]{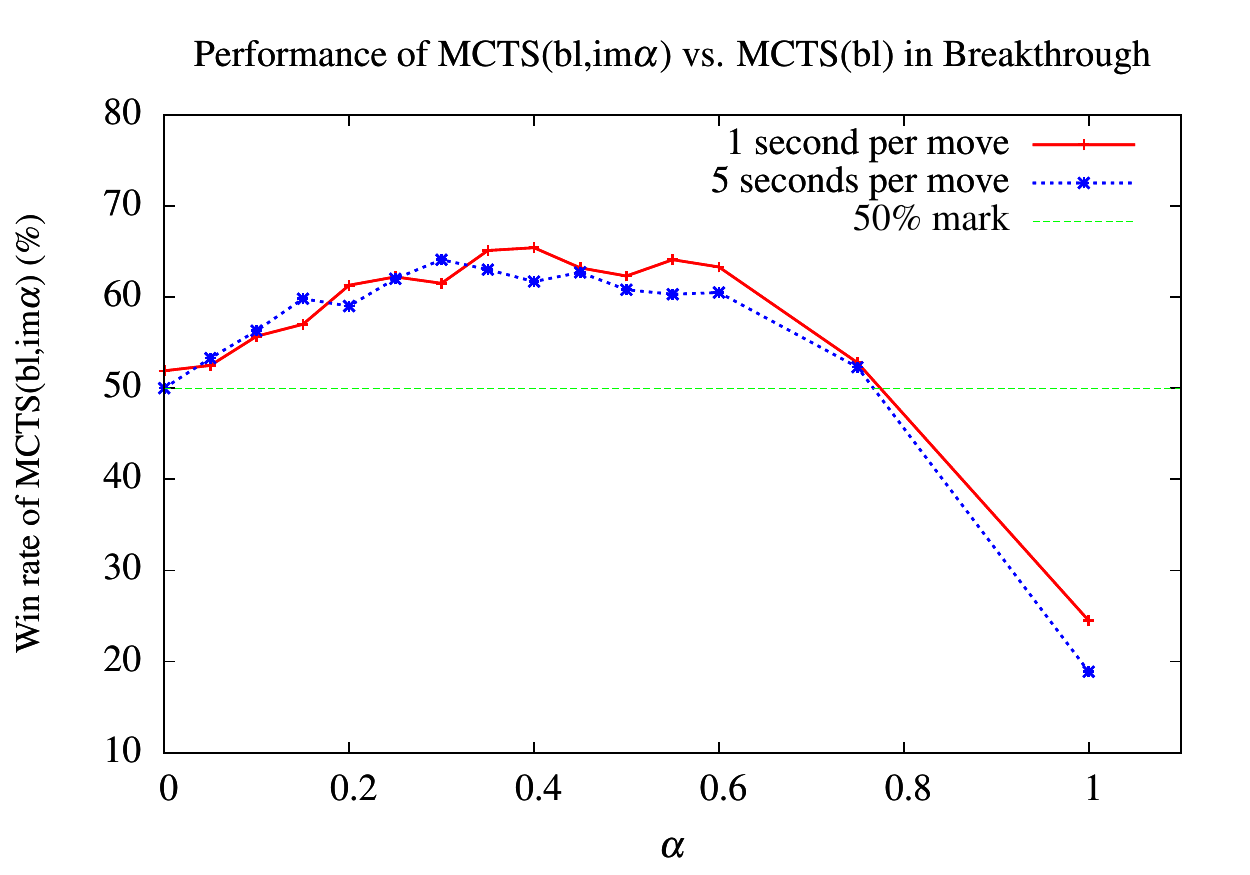}\\
\includegraphics[scale=0.7]{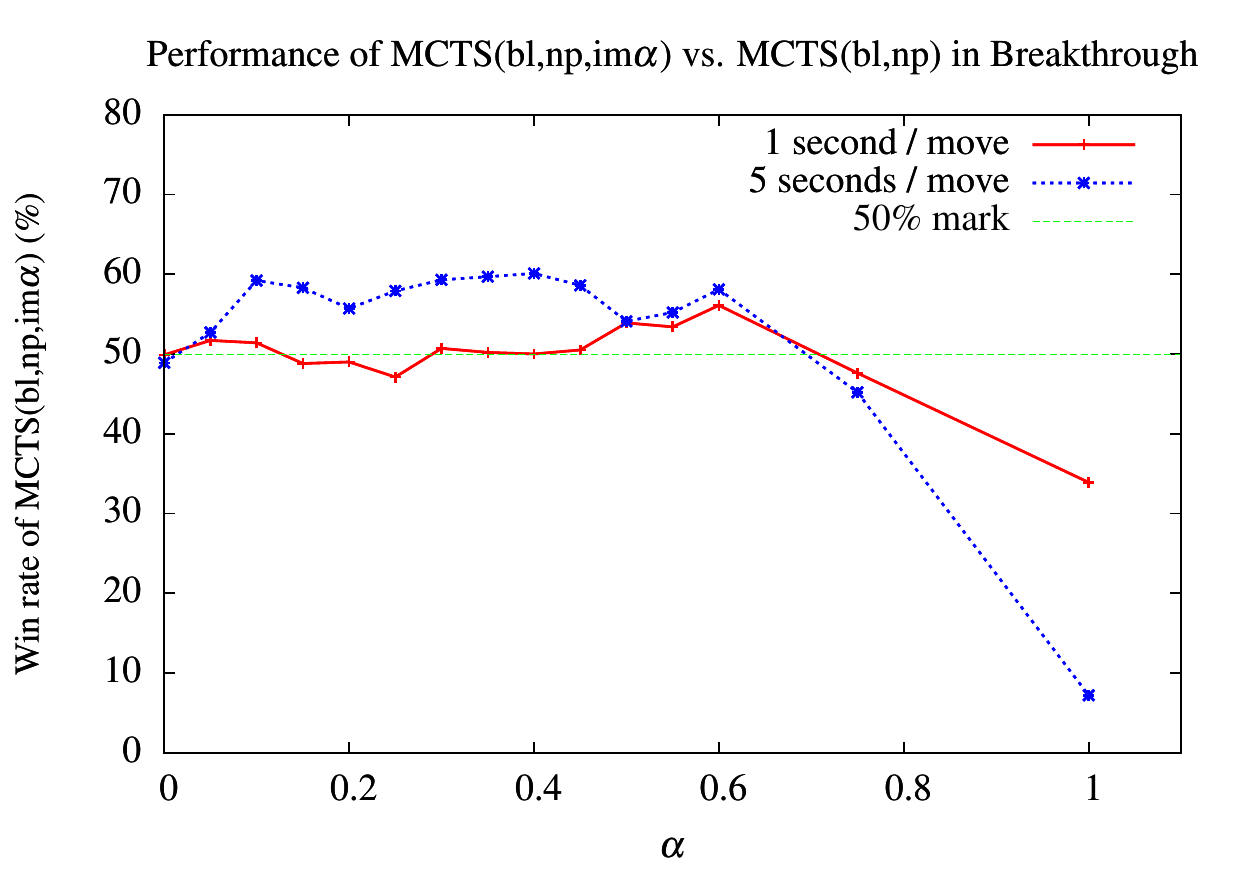}
\caption{Results in Breakthrough against 
baseline player MCTS(ege$0.1$,det$0.5$).  
Each point represents 1000 games. The top graph excludes node priors, bottom graph includes node priors.} 
\label{fig:bt-base-alpha}
\end{center}
\end{figure}

Our first set of experiments uses the simple evaluation function, efMS. At the end of this subsection, we 
include experiments for the sophisticated evaluation function efLH. 

We first determined the best playout strategy amongst fixed and dynamic early 
terminations and $\epsilon$-greedy playouts.
Our best fixed early terminations player was fet$20$ and best $\epsilon$-greedy player was ege$0.1$.
Through systematic testing on 1000 games per pairing, we determined that the best playout 
policy when using efMS is the combination (ege$0.1$,det$0.5$). 
The detailed test results are found in Appendix B.
To ensure that this combination of early termination strategies is indeed superior to just the improved 
playout policy on its own, we also played MCTS(ege$0.1$,det$0.5$) against MCTS(ipp). 
MCTS(ege$0.1$,det$0.5$) won 68.8\% of these games.
MCTS(ege$0.1$,det$0.5$) is the best baseline player that we could produce given three separate 
parameter-tuning tournaments, for all the playout enhancements we have tried using efMS, over 
thousands of played games. Hence, we use it as our primary benchmark for comparison in the 
rest of our experiments with efMS. 
For convenience, we abbreviate this baseline player (MCTS(ege$0.1$,det$0.5$)) to MCTS(bl). 

We then played MCTS with implicit minimax backups, MCTS(bl,im$\alpha$), against MCTS(bl) for a 
variety different values for $\alpha$. The results are shown in the top of Figure~\ref{fig:bt-base-alpha}.
Implicit minimax backups give an advantage for $\alpha \in [0.1,0.6]$ under both one- and five-second 
search times. 
When $\alpha > 0.6$, MCTS(bl,im$\alpha$) acts like greedy best-first minimax.
To verify that the benefit was not only due to the optimized playout policy, we performed two experiments. 
First, we played MCTS without playout terminations, MCTS(ipp,im$0.4$) against MCTS(ipp). 
MCTS(ipp,im$0.4$) won 82.3\% of these games. 
We then tried giving both players fixed early terminations, and played MCTS(ipp,fet$20$,im$0.4$) versus 
MCTS(ipp,fet$20$). MCTS(ipp,fet$20$,im$0.4$) won 87.2\% of these games.

 
The next question was whether the mixing static evaluation values themselves ($v_0(s)$) at node $s$ 
was the source of the benefit or whether the minimax backup values ($v^{\tau}_s$) were the contributing factor.
Therefore, we tried MCTS(bl, im$0.4$) against a baseline player that uses constant bias over the static 
evaluations, \ie uses 
\[ \hQ^{CB}(s,a) = (1-\alpha)Q + \alpha v_0(s'), \mbox{ where } s' = \cT(s,a),\]
and also against a player using a progressive bias of the implicit minimax values, \ie 
\[ \hQ^{PB}(s,a) = (1-\alpha)Q + \alpha v^{\tau}_{s,a}/(n_{s,a} + 1), \]
with $\alpha = 0.4$ in both cases.
MCTS(bl,im$0.4$) won 67.8\% against MCTS(bl,$\hQ^{CB}$). 
MCTS(bl,im$0.4$) won 65.5\% against MCTS(bl,$\hQ^{PB}$). 
A different decay function for the weight placed on $v^{\tau}_s$ could further improve 
the advantage of implicit minimax backups. We leave this as a topic for future work. 

\begin{table}[t]
\begin{center}
\begin{tabular}{|c|c|c|c|}
\hline
{\bf Player A}              & {\bf Player B}         & {\bf A Wins} (\%)  \\
\hline
MCTS(ipp)                   & MCTS(random playouts)  & 94.30 $\pm$ 1.44    \\
\hline 
\hline 
\multicolumn{3}{|c|}{Experiments using only efMS}  \\ 
\hline
MCTS(ege$0.1$,det$0.5$)     & MCTS(ipp)              & 68.80 $\pm$ 2.88    \\
MCTS(ipp,im$0.4$)           & MCTS(ipp)              & 82.30 $\pm$ 2.37    \\ 
MCTS(ipp,fet$20$,im$0.4$)   & MCTS(ipp,fet$20$)      & 87.20 $\pm$ 2.07    \\
MCTS(bl,im$0.4$)            & MCTS(bl,$\hQ^{CB}$)    & 67.80 $\pm$ 2.90    \\
MCTS(bl,im$0.4$)            & MCTS(bl,$\hQ^{PB}$)    & 65.50 $\pm$ 2.95    \\
MCTS(bl,im$0.6$)            & MCTS(bl)               & 63.30 $\pm$ 2.99    \\
MCTS(bl,im$0.6$,np)         & MCTS(bl)               & 77.90 $\pm$ 2.57    \\
\hline
\hline
\multicolumn{3}{|c|}{Experiments using efMS and efLH}  \\ 
\hline
MCTS(efMS,bl)               & MCTS(efLH,bl')         & 40.20 $\pm$ 3.04    \\
MCTS(efMS,bl,np)            & MCTS(efLH,bl')         & 78.00 $\pm$ 2.57    \\
MCTS(efMS,bl,np,im$0.4$)    & MCTS(efLH,bl')         & 84.90 $\pm$ 2.22    \\ 
MCTS(efMS,bl,im$0.4$)       & MCTS(efLH,bl',im$0.6$) & 53.40 $\pm$ 2.19    \\
\hline
\end{tabular}
\end{center}
\caption{Summary of results in Breakthrough, with 95\% confidence intervals.}
\label{tbl:btsummary}
\end{table}

We then evaluated MCTS(im$0.4$) against {\it maximum backpropagation} proposed as an alternative 
backpropagation in the original MCTS work~\cite{Coulom06Efficient}. This enhancement
modifies line 24 of the algorithm to the following: 
\[
\mathbf{if}~n_{s} \ge T~\mathbf{then~return}~\max_{a \in \cA(s)} \hQ(s,a)~\mathbf{else~return}~r.
\]
The results for several values of $T$ are given in Table~\ref{tbl:maxbackprop}. 

\begin{table}[t]
\begin{center}
\begin{tabular}{|c|ccccccc|}
\hline
                     & \multicolumn{7}{|c|}{$T$ (in thousands)}             \\
Time                & 0.1  & 0.5  & 1    & 5    & 10   & 20 & 30 \\
\hline
1s                 & 81.9 & 73.1 & 69.1 & 65.2 & 63.6 & 66.2 & 67.0 \\
\hline
\end{tabular}
\end{center}
\caption{Win rates (\%) of MCTS(bl,im$0.4$) vs. max backpropagation in Breakthrough, 
for $T \in \{ 100, \cdots, 30000 \}$. }
\label{tbl:maxbackprop}
\end{table}

Another question is whether to prefer implicit minimax backups over {\it node priors} (abbreviated np)~\cite{Gelly07Combining}, 
which initializes each new leaf node with wins and losses based on prior knowledge. Node priors were first used in Go, 
and have also used in path planning problems~\cite{Eyerich10High}.
We use the scheme that worked well in~\cite{Lorentz13Breakthrough}
which takes into account the safety of surrounding pieces, and scales the counts by the 
time setting (10 for one second, 50 for five seconds).
We ran an experiment against the baseline player with node priors enabled,
MCTS(bl,im$\alpha$,np) versus MCTS(bl,np). 
The results are shown at the bottom of Figure~\ref{fig:bt-base-alpha}. When combined at one second of 
search time, implicit minimax backups still seem to give an advantage for $\alpha \in [0.5,0.6]$, and at five 
seconds gives an advantage for $\alpha \in [0.1,0.6]$. To verify that the combination is complementary,
we played MCTS(bl,im$0.6$) with and without node priors each against the baseline player. The player with
node priors won 77.9\% and the one without won 63.3\%.


A summary of these comparisons is given in Table~\ref{tbl:btsummary}.\\

\noindent{\it MCTS Using Lorentz \& Horey Evaluation Function}

\vspace{0.2cm} 

We now run experiments using the more sophisticated evaluation function from~\cite{Lorentz13Breakthrough}, 
efLH, that assigns specific piece count values depending on their position on the board. 
Rather than repeating all of the above experiments, we chose simply to compare baselines and to repeat
the initial experiment, all using 1 second of search time.

\begin{figure}[t]
\begin{center}
\includegraphics[scale=0.68]{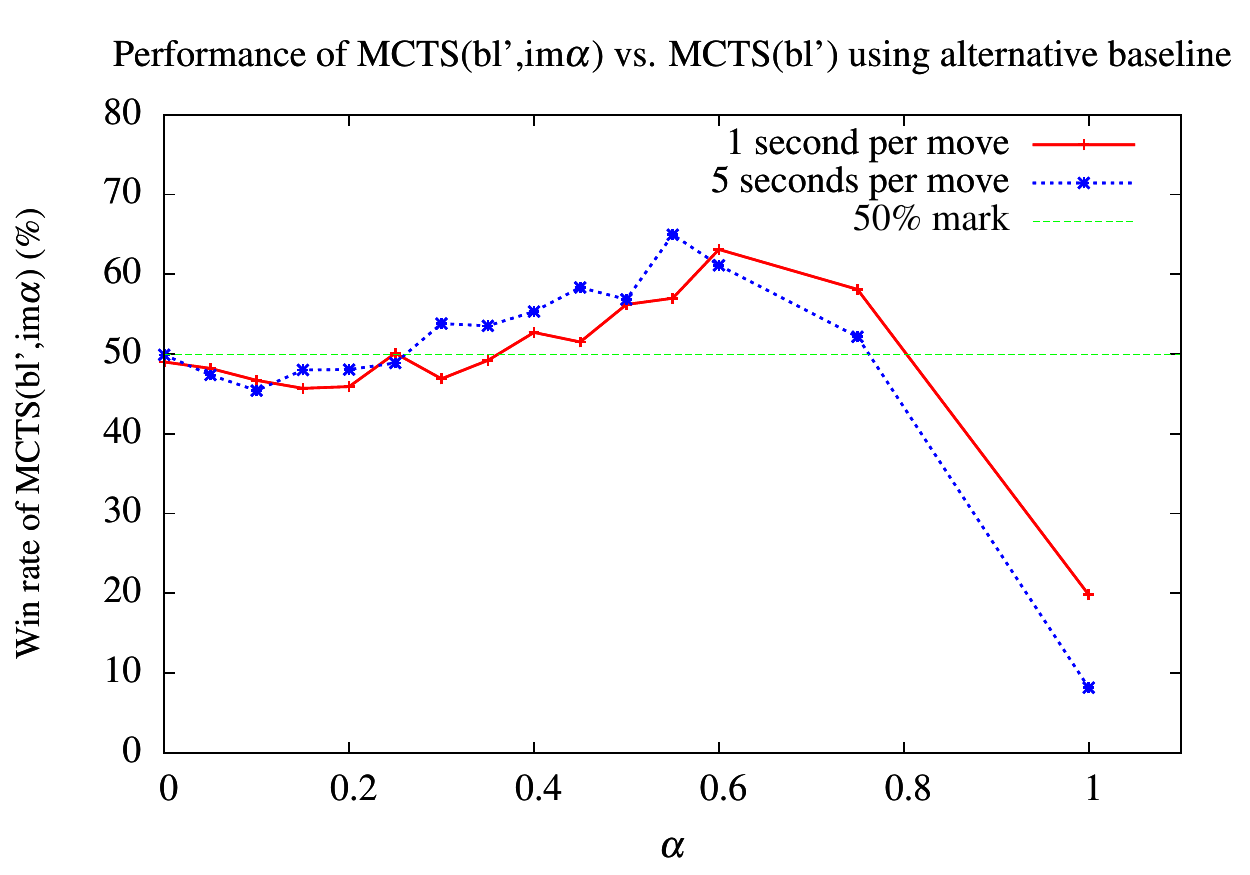}
\caption{Results of varying $\alpha$ in Breakthrough using the alternative baseline player.
Each point represents 1000 games.} 
\label{fig:bt-alt-alpha}
\end{center}
\end{figure}

The best playout with this evaluation function is fet$20$ with node priors, which we call the alternative baseline, 
abbreviated bl'.
That is, we abbreviate MCTS(ipp,fet$20$,np) to MCTS(bl').
We rerun the initial $\alpha$ experiment using the alternative baseline, 
which uses the Lorentz \& Horey evaluation function, to find the best implicit minimax player using this 
more sophisticated evaluation function. Results are shown in Figure~\ref{fig:bt-alt-alpha}. 
In this case the best range is $\alpha \in [0.5,0.6]$ for one second and $\alpha \in [0.5,0.6]$ for
five seconds.
We label the best player in this figure using the alternative baseline MCTS(efLH,bl',im$0.6$). 

In an effort to explain the relative strengths of each evaluation function, we then compared the two baseline players. 
Our baseline MCTS player, MCTS(efMS,bl), wins 40.2\% of games against the alternative baseline, MCTS(efLH,bl'). 
When we add node priors, MCTS(efMS,bl,np) wins 78.0\% of games against MCTS(efLH,bl'). 
When we also add implicit minimax backups ($\alpha = 0.4$), the win rate of MCTS(efMS,bl,im$0.4$,np) versus MCTS(efLH,bl') rises again to 84.9\%. 
Implicit minimax backups improves performance against a stronger benchmark player, even when using a simpler evaluation function. 

We then played 2000 games of the two best players for the respective evaluation functions against each other, 
that is we played MCTS(efMS,bl,np,im$0.4$) against MCTS(efLH,bl',im$0.6$). 
MCTS(efMS,bl,np,im$0.4$) wins 53.40\% of games. 
Given these results, it could be that a more defensive and less granular evaluation function is preferred 
in Breakthrough when given only 1 second of search time. 
The results in our comparison to $\alpha\beta$ in the next subsection seem to suggest this as well.\\

\noindent {\it Comparison to $\alpha\beta$ Search} 

\vspace{0.2cm} 

\begin{table}[t]
\begin{center}
\begin{tabular}{ccccc|c}
Ev. Func.  & Player              & Opp.                 & $n$    & $t$ (s) & Res. (\%) \\
\hline
(Both)     & $\alpha\beta$(efMS) & $\alpha\beta$(efLH)  & 2000   & 1       & 70.40     \\
(Both)     & $\alpha\beta$(efMS) & $\alpha\beta$(efLH)  &  500   & 5       & 53.40     \\
(Both)     & $\alpha\beta$(efMS) & $\alpha\beta$(efLH)  &  400   & 10      & 31.25     \\
\hline
\hline
efMS       & MCTS(bl)              & $\alpha\beta$    & 2000   & 1       & 27.55     \\     
efMS       & MCTS(bl)              & $\alpha\beta$    & 1000   & 5       & 39.00     \\     
efMS       & MCTS(bl)              & $\alpha\beta$    &  500   & 10      & 47.60     \\     
\hline
efMS       & MCTS(bl,im$0.4$)     & $\alpha\beta$    & 2000   & 1       & 45.05     \\     
efMS       & MCTS(bl,im$0.4$)     & $\alpha\beta$    & 1000   & 5       & 61.60     \\     
efMS       & MCTS(bl,im$0.4$)     & $\alpha\beta$    &  500   & 10      & 61.80     \\     
\hline
\hline
efLH       & MCTS(bl')              & $\alpha\beta$    & 2000   & 1       &  7.90      \\     
efLH       & MCTS(bl')              & $\alpha\beta$    & 1000   & 5       & 10.80      \\     
efLH       & MCTS(bl')              & $\alpha\beta$    &  500   & 10      & 12.60      \\     
efLH       & MCTS(bl')              & $\alpha\beta$    &  500   & 20      & 18.80      \\     
efLH       & MCTS(bl')              & $\alpha\beta$    &  500   & 30      & 19.40      \\     
efLH       & MCTS(bl')              & $\alpha\beta$    &  500   & 60      & 24.95      \\     
efLH       & MCTS(bl')              & $\alpha\beta$    &  130   & 120     & 25.38      \\     
\hline
efLH       & MCTS(bl',im$0.6$)     & $\alpha\beta$    & 2000   & 1       & 28.95      \\     
efLH       & MCTS(bl',im$0.6$)     & $\alpha\beta$    & 1000   & 5       & 39.30      \\     
efLH       & MCTS(bl',im$0.6$)     & $\alpha\beta$    &  500   & 10      & 41.20      \\     
efLH       & MCTS(bl',im$0.6$)     & $\alpha\beta$    &  500   & 20      & 45.80      \\     
efLH       & MCTS(bl',im$0.6$)     & $\alpha\beta$    &  500   & 30      & 46.20      \\     
efLH       & MCTS(bl',im$0.6$)     & $\alpha\beta$    &  500   & 60      & 55.60      \\     
efLH       & MCTS(bl',im$0.6$)     & $\alpha\beta$    &  130   & 120     & 61.54      \\     
\hline
\end{tabular}
\end{center}
\caption{Summary of results versus $\alpha\beta$. 
Here, $n$ represents the number of games played and $t$ time in seconds per search.
Win rates are for the Player (in the left column).}
\label{tbl:ab_vs_mcts}
\end{table}

A natural question is how MCTS with implicit minimax backups compares to $\alpha\beta$ search. 
So, here we compare MCTS with implicit minimax backups versus $\alpha\beta$ search.
Our $\alpha\beta$ search player uses iterative deepening and a static move ordering. 
The static move ordering is based on the same information used in the improved playout
policies: decisive and anti-decisive moves are first, then captures of defenseless pieces, 
then all other captures, and finally regular moves.
The results are listed in Table~\ref{tbl:ab_vs_mcts}.

The first observation is that the performance of MCTS (vs. $\alpha\beta$) 
increases as search time increases. This is true in all cases, using either evaluation function, 
with and without implicit minimax backups. This is similar to observations in 
Lines of Action~\cite{Winands11AB} and multiplayer MCTS~\cite{Sturtevant08An,Nijssen13}.

The second observation is that MCTS(im$\alpha$) performs significantly better 
against $\alpha\beta$ than the baseline player at the same search time. 
Using efMS in Breakthrough with 5 seconds of search time, MCTS(im$0.4$) performs significantly 
better than both the baseline MCTS player and $\alpha\beta$ search on their own. 

The third observation is that MCTS(im$\alpha$) benefits significantly from weak 
heuristic information, more so than $\alpha\beta$. 
When using efMS, MCTS takes less long to do better against $\alpha\beta$, 
possibly because MCTS makes better use of weaker information. 
When using efLH, $\alpha\beta$ preforms significantly better against MCTS at low time settings.  
However, it unclear whether this due to $\alpha\beta$ improving or MCTS worsening.
Therefore, we also include a comparison of the $\alpha\beta$ players
using efMS versus efLH. What we see is that at 1 second, efMS benefits $\alpha\beta$ more, but as time 
increases efLH seems to be preferred. 
Nonetheless, when using efLH, there still seems to be a point where, if given enough search time
the performance of MCTS(im$0.6$) surpasses that of $\alpha\beta$. 

\subsection{Lines of Action}

In subsection~\ref{sec:bt}, we compared the performance of MCTS(im$\alpha$) to a basic $\alpha\beta$
search player. Our main question at this point is how MCTS(im$\alpha$) could perform in a game with 
stronger play due to using proven enhancements in both $\alpha\beta$ and MCTS.
For this analysis, we now consider the well-studied game Lines of Action (LOA). 

LOA is a turn-taking alternating-move game played on an 8-by-8 board that uses checkers board and pieces.
The goal is to connect all your pieces into a single connected group (of any size), 
where the pieces are connected via adjacent and diagonals squares. A piece may move in any direction, but the number of squares 
it may move depends on the total number of pieces in the line, including opponent pieces. A piece may jump over its own
pieces but not opponent pieces. Captures occur by landing on opponent pieces. 

The MCTS player is MC-LOA, whose implementation and enhancements are described in \cite{Winands10MCTS-LOA}. 
MC-LOA is a world-champion engine winning the latest Olympiad. The benchmark $\alpha\beta$ player is MIA, the world-best $\alpha\beta$-player 
upon which MC-LOA is based, winning 4 Olympiads. MC-LOA uses MCTS-Solver, progressive bias, and highly-optimized $\alpha\beta$
playouts. MIA includes the following enhancements: static move ordering, iterative deepening, killer moves, history heuristic, enhanced 
transposition table cutoffs, null-move pruning, multi-cut, realization probability search, quiescence search, and negascout/PVS. 
The evaluation function used is the used in MIA~\cite{Winands06MIA}.
All of the results in LOA are based 100 opening board 
positions.\footnote{\small https://dke.maastrichtuniversity.nl/m.winands/loa/} 

\begin{figure}[t]
\begin{center}
\includegraphics[scale=0.7]{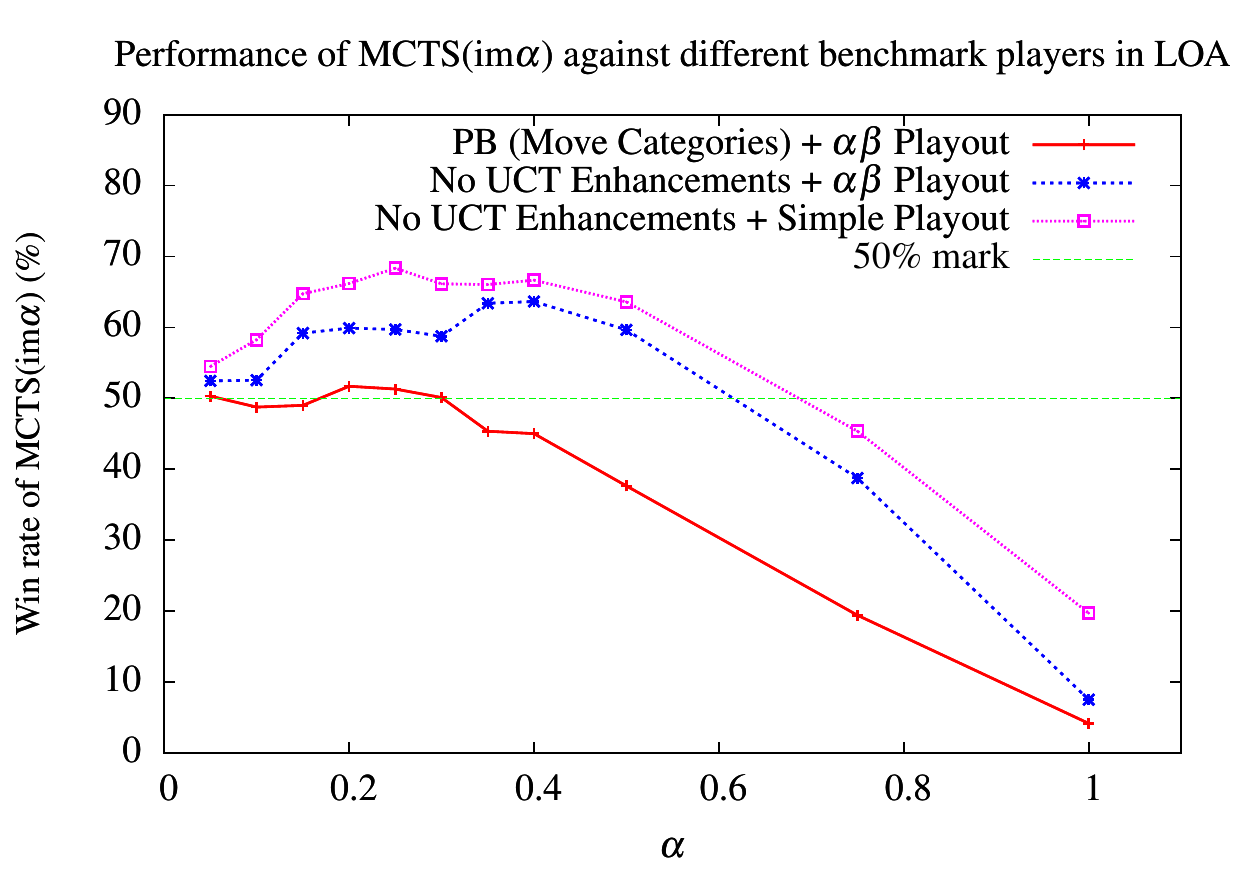}
\caption{Results in LOA. Each data point represents 1000 games with 1 second of search time. } 
\label{fig:loa-alpha}
\end{center}
\end{figure}

\begin{table}[t]
\begin{center}
\begin{tabular}{ccccc|c}
Options & Player        & Opp. & $n$ & $t$ & Res. (\%) \\
\hline
\hline
PB      & MCTS(im$\alpha$) & MCTS     & 32000 & 1        & 50.59       \\
PB      & MCTS(im$\alpha$) & MCTS     & 6000  & 5        & 50.91       \\
\hline
$\neg$PB & MCTS(im$\alpha$) & MCTS & 1000 & 1  & 59.90       \\
$\neg$PB & MCTS(im$\alpha$) & MCTS & 6000 & 5  & 63.10       \\
$\neg$PB & MCTS(im$\alpha$) & MCTS & 2600 & 10 & 63.80       \\
\hline
$\neg$PB & MCTS             & $\alpha \beta$ & 2000 & 5  & 40.0       \\
$\neg$PB & MCTS(im$\alpha$) & $\alpha \beta$ & 2000 & 5 & 51.0       \\
\hline
PB       & MCTS             & $\alpha \beta$ & 20000 & 5  & 61.8       \\
PB       & MCTS(im$\alpha$) & $\alpha \beta$ & 20000 & 5  & 63.3       \\
\hline
\end{tabular}
\end{center}
\caption{Summary of results for players and opponent pairings in LOA. 
All MCTS players use $\alpha \beta$ playouts and MCTS(im$\alpha$) players use $\alpha = 0.2$. 
Here, $n$ represents the number of games played and $t$ time in seconds per search.}
\label{tbl:loaresults}
\end{table}

We repeat the implicit minimax backups experiment with varying $\alpha$. At first, we use standard UCT without enhancements 
and a simple playout that is selects moves non-uniformly at random based on the move categories, and uses the early cut-off strategy. 
Then, we enable shallow $\alpha\beta$ searches in the playouts described in~\cite{Winands11AB}. 
Finally, we enable the progressive bias based on move categories in addition to the $\alpha\beta$ playouts. The results for these 
three different settings are shown in Figure~\ref{fig:loa-alpha}. As before, we notice that in the first two situations,
implicit minimax backups with $\alpha \in [0.1,0.5]$ can lead to better performance. When the progressive bias based on move 
categories is added, the advantage diminishes. However, we do notice that $\alpha \in [0.05,0.3]$ seems to not significantly 
decrease the performance. 

Additional results are summarized in Table~\ref{tbl:loaresults}. From the graph, we reran $\alpha = 0.2$ with progressive bias for 
32000 games giving a statistically significant (95\% confidence) win rate of 50.59\%. 
We also tried increasing the search time, in both cases (with and without progressive bias), 
and observed a gain in performance at five and ten seconds. 
In the past, the strongest LOA player was MIA, which was based on $\alpha \beta$ search. Therefore, we also test our MCTS with 
implicit minimax backups against an $\alpha \beta$ player based on MIA. When progressive bias is disabled, implicit minimax backups
increases the performance by 11 percentage points. There is also a small increase in performance when progressive bias is enabled. 
Also, at $\alpha = 0.2$, it seems that there is no statistically significant case of implicit minimax backups hurting performance. 

\subsection{Discussion: Traps and Limitations} 

The initial motivation for this work was driven by the trap moves, which pose problems in 
MCTS~\cite{Ramanujan11Tradeoffs,Baier13MinimaxHybrids,Gudmundsson13Sufficiency}. 
However, in LOA we observed that implicit minimax backups did not speed up MCTS when 
solving a test set of end game positions.
We tried to construct an example board in Breakthrough 
to demonstrate how implicit minimax backups deals with problems with traps. We were unable to do so. In our experience, 
traps are effectively handled by the improved playout policy. Even without early terminations, 
simply having decisive and anti-decisive moves and preferring good capture moves seems to be enough to handle traps in Breakthrough.
Also, even with random playouts, an efficient implementation with MCTS-Solver handles shallow traps.
Therefore, we believe that the explanation for the advantage offered by implicit minimax backups is more subtle
than simply detecting and handling traps. In watching several Breakthrough games, it seems that MCTS with implicit minimax backups 
builds ``fortress'' structures \cite{Guid12Fortress} that are then handled better than standard MCTS.  

While we have shown positive results in a number of domains, we recognize that this 
technique is not universally applicable. We believe that implicit minimax backups work because there is short-term tactical 
information, which 
is not captured in the long-term playouts, but is captured by the implicit minimax procedure. Additionally, we suspect that 
there must be strategic 
information in the playouts which is not captured in the shallower minimax backups. Thus, success depends on both the domain and 
the evaluation function used. 
We also ran experiments for implicit minimax backups in Chinese Checkers and the card game Hearts, 
and there was no significant improvement in performance,  
but more work has to be performed to understand if we would find success with a better evaluation function.

\section{Conclusion}

We have introduced a new technique called implicit minimax backups for MCTS. 
This technique stores the information from both sources separately, only combining the
two sources to guide selection. Implicit minimax can lead to stronger play even with
simple evaluation functions, which are often readily available. 
In Breakthrough, our evaluation shows that implicit minimax backups increases 
the strength of MCTS significantly compared to similar techniques for improving MCTS
using domain knowledge. Furthermore, the technique improves 
performance in LOA, a more complex domain with sophisticated knowledge and 
strong MCTS and $\alpha \beta$ players. 
The range $\alpha \in [0.15,0.4]$ seems to be a safe choice.
In Breakthrough, this range is higher, $[0.5,0.6]$, 
when using node priors at lower time settings and when using the alternative baseline. 

For future work, we would like to apply the technique in other games, such as Amazons, 
and plan to investigate improving initial evaluations $v_0(s)$ using quiescence search.
We hope to compare or combine implicit minimax backups to/with other 
minimax hybrids from~\cite{Baier13MinimaxHybrids}. 
Differences between $v^{\tau}_{s,a}$ and $Q(s,a)$ could indicate parts of the tree that require 
more search and hence help guide selection. 
Parameters could be modified online. For example, $\alpha$ could be changed based on the outcomes
of each choice made during the game, and $Q(s,a)$ could be used for online search 
bootstrapping of evaluation function weights~\cite{Veness09Bootstrapping}.
Finally, the technique could also work in general game-playing using learned
evaluation functions~\cite{Finnsson10Learning}. \\

{\small
\noindent {\bf Acknowledgments.} This work is partially funded by the Netherlands 
Organisation for Scientific Research (NWO) in the framework of the project Go4Nature, grant number 612.000.938. 
}

%
%

\bibliography{im-mcts}
\bibliographystyle{plain}

\vspace{20cm}

\appendix

\section{Game Images}
\label{app:images}

This appendix shows images of each game to help visualize the objectives.


\begin{figure}[h!]
\begin{center}
\includegraphics[scale=0.21]{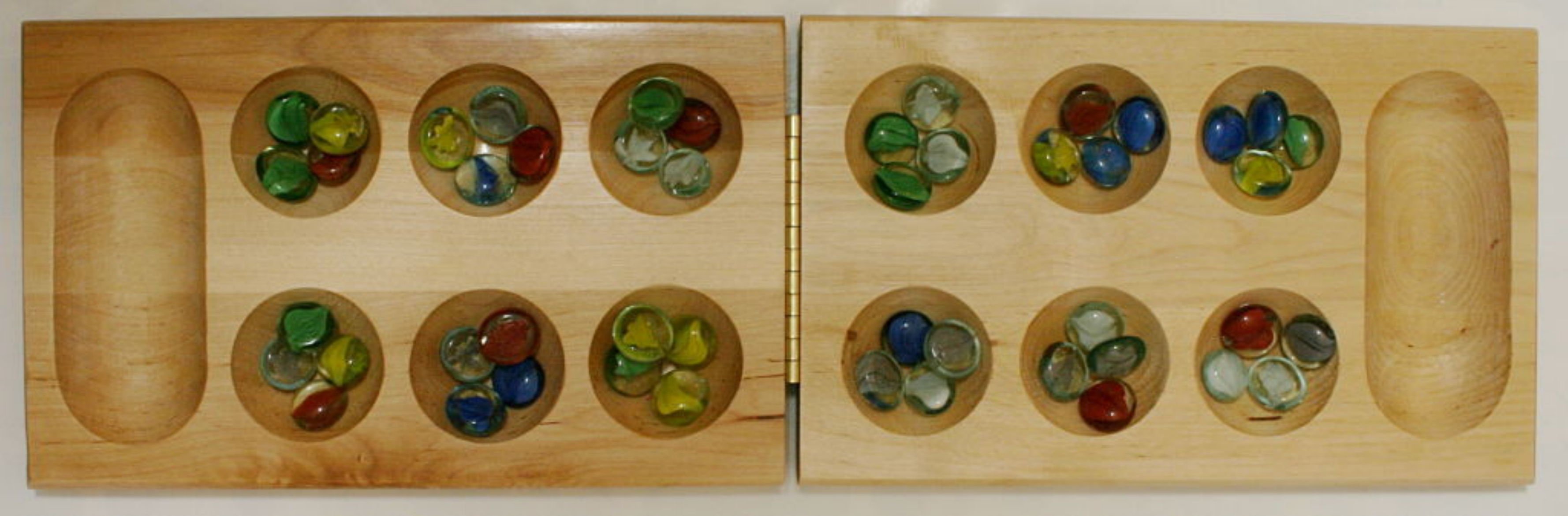}
\end{center}
\caption{An example position in Kalah (with only 4 stones in each house.) 
Retrieved June 1st.
Image source: http://upload.wikimedia.org/wikipedia/commons/6/6d/Wooden\_Mancala\_board.jpg. \label{fig:kalah-image}}
\end{figure}

\begin{figure}[h!]
\begin{center}
\includegraphics[scale=0.4]{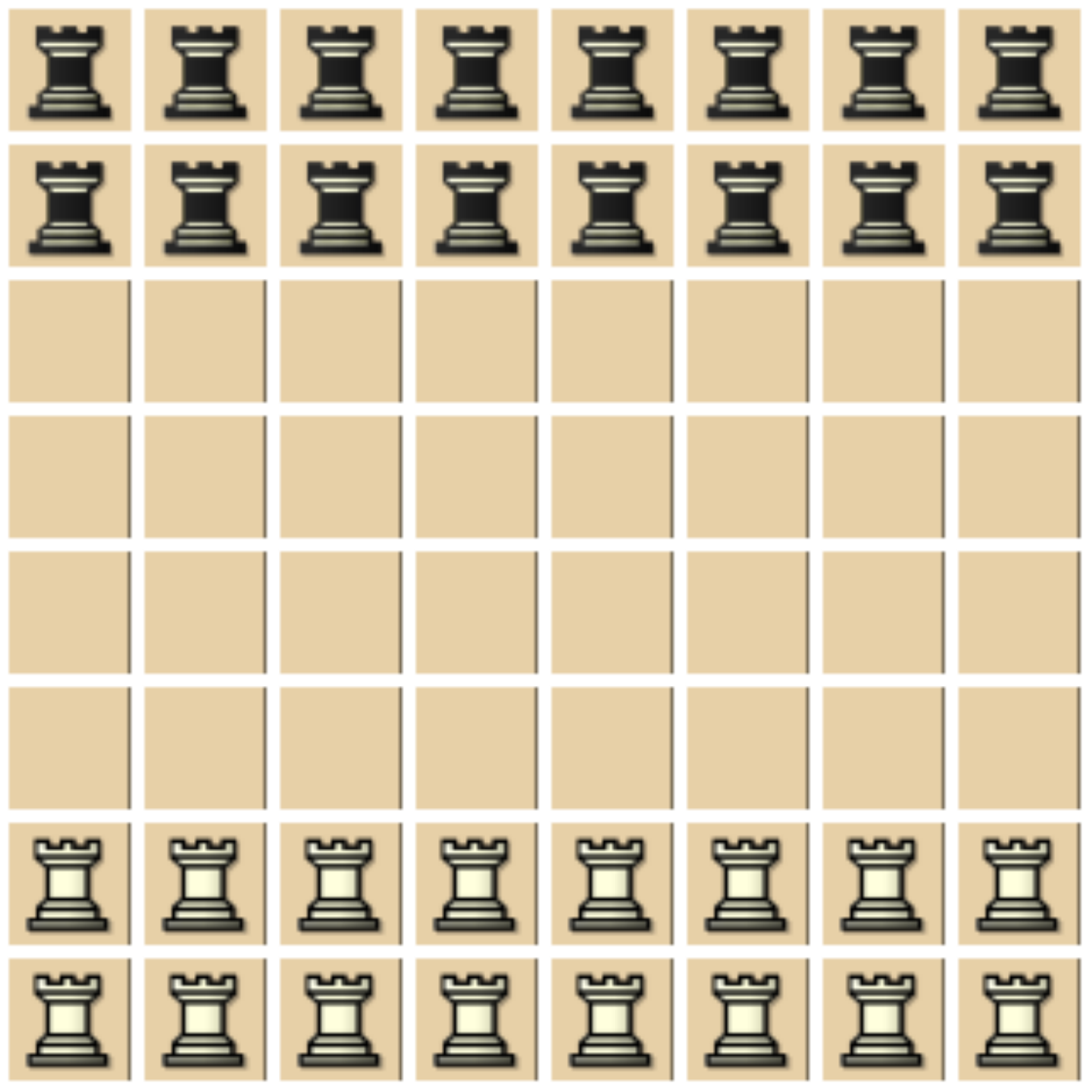}
\end{center}
\caption{Breakthrough initial position, as depicted on the online turn-based game web site Little Golem. 
Retrieved June 1st, 2014. 
Image source: http://www.littlegolem.net/jsp/games/break01.png. 
\label{fig:bt-image}}
\end{figure}

\begin{figure}[h!]
\begin{center}
\includegraphics[scale=0.4]{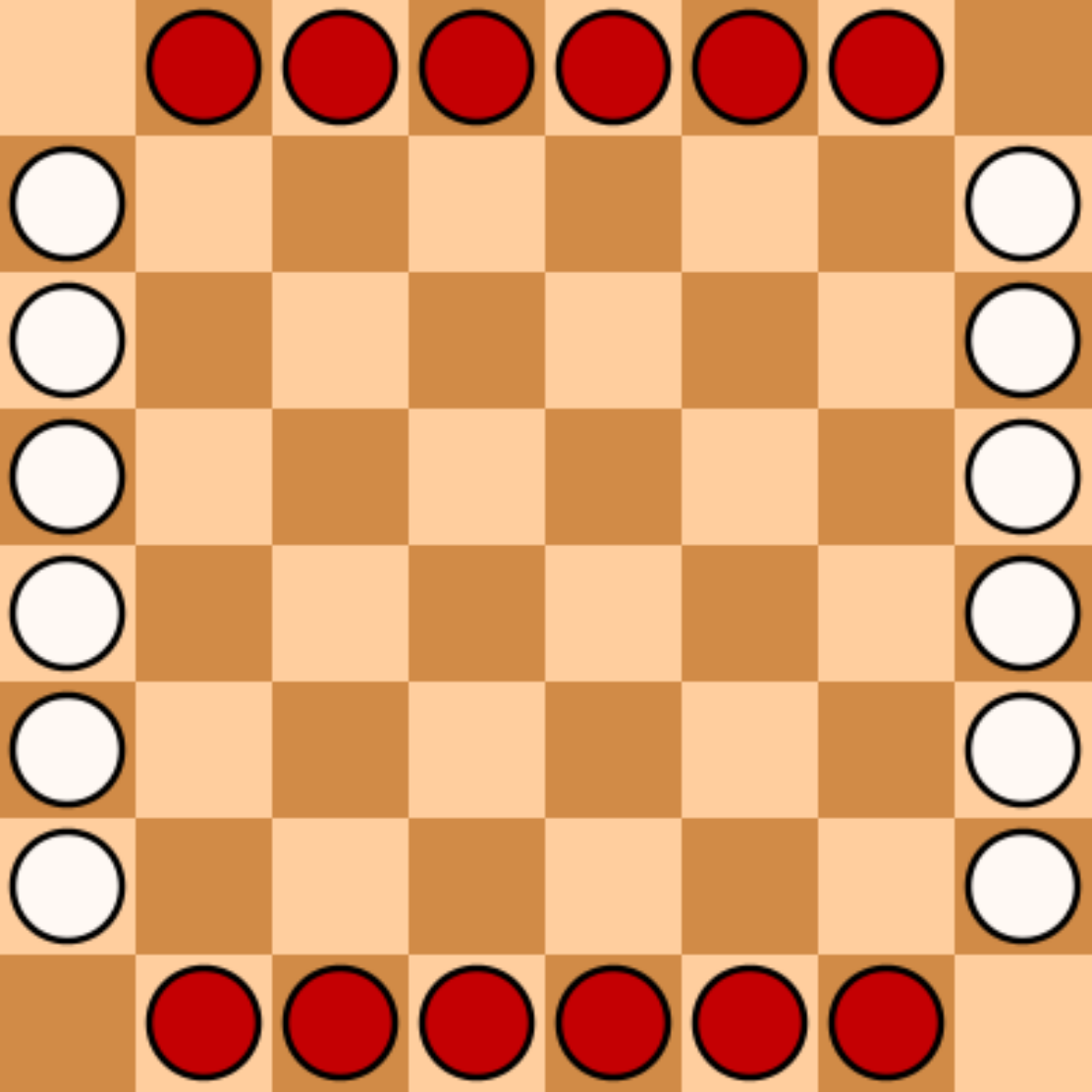}
\end{center}
\caption{Starting position of Lines of Action. 
Retrieved June 1st, 2014. 
Image source: http://upload.wikimedia.org/wikipedia/commons/1/1b/Lines\_of\_Action.svg. 
\label{fig:loa-image}}
\end{figure}

\section{Tournaments and Playout Comparisons}
\label{app:results}

This appendix includes details of the results of played games to determine the best baseline players. 

\subsection{Parameter Values for Breakthrough and Kalah}

Table~\ref{tbl:parmvalues} shows the parameter values that were used in parameter-tuning experiments.

\begin{table}[h!]
\begin{center}
\begin{tabular}{|l|l|}
\hline
Technique & Parameter set \\
\hline
fet$x$          & $\{ 0, 1, \ldots, 5, 8, 10, 12, 16, 20, 30, 50, 100, 1000 \}$ \\
det$x$         & $\{ .1, .2, .3, , .4, .5, .55, .6, .65, .7, .75, .8, .85, .9 \}$ \\
ege$\epsilon$  & $\{ 0, .05, .1, .15, .2, .3, .4, .5, .6, .7, .8, .9, 1 \}$ \\
im$\alpha$     & $\{ 0, .05, .1, .15, \ldots, .55, .6, .75, 1 \}$ \\
\hline
\end{tabular}
\end{center}
\caption{Parameter value sets. \label{tbl:parmvalues}}
\label{tbl:parmsets}
\end{table}

\subsection{Kalah Playout Optimization}

In Kalah, each matchup included 1000 games, but as mentioned in the main part of the 
paper, only the wins and losses are shown and unbalanced boards are removed. 

\subsubsection{Fixed Early Termination Tournament}

\begin{verbatim}
round 1
winner mcts_h_fet0 (368) vs. loser mcts_h_fet1000 (61)
winner mcts_h_fet1 (408) vs. loser mcts_h_fet100 (61)
winner mcts_h_fet2 (458) vs. loser mcts_h_fet50 (61)
winner mcts_h_fet3 (460) vs. loser mcts_h_fet30 (37)
winner mcts_h_fet4 (429) vs. loser mcts_h_fet20 (44)
winner mcts_h_fet5 (223) vs. loser mcts_h_fet10 (83)
mcts_h_fet8 gets a by

round 2
winner mcts_h_fet0 (181) vs. loser mcts_h_fet8 (169)
winner mcts_h_fet5 (189) vs. loser mcts_h_fet1 (116)
winner mcts_h_fet4 (166) vs. loser mcts_h_fet2 (115)
mcts_h_fet3 gets a by

round 3
winner mcts_h_fet3 (161) vs. loser mcts_h_fet0 (124)
winner mcts_h_fet4 (132) vs. loser mcts_h_fet5 (122)

round 4
winner mcts_h_fet4 (139) vs. loser mcts_h_fet3 (110)

Winner: mcts_h_fet4
\end{verbatim}

\subsubsection{Epsilon-greedy Playout Tournament}

\begin{verbatim}
round 1
winner mcts_h_ege1.0 (232) vs. loser mcts_h_ege0.0 (205)
winner mcts_h_ege0.9 (224) vs. loser mcts_h_ege0.05 (210)
winner mcts_h_ege0.1 (207) vs. loser mcts_h_ege0.8 (195)
winner mcts_h_ege0.15 (219) vs. loser mcts_h_ege0.7 (213)
winner mcts_h_ege0.2 (244) vs. loser mcts_h_ege0.6 (211)
winner mcts_h_ege0.3 (233) vs. loser mcts_h_ege0.5 (197)
mcts_h_ege0.4 gets a by

round 2
winner mcts_h_ege1.0 (276) vs. loser mcts_h_ege0.4 (173)
winner mcts_h_ege0.9 (231) vs. loser mcts_h_ege0.3 (195)
winner mcts_h_ege0.1 (212) vs. loser mcts_h_ege0.2 (196)
mcts_h_ege0.15 gets a by

round 3
winner mcts_h_ege1.0 (218) vs. loser mcts_h_ege0.15 (206)
winner mcts_h_ege0.9 (206) vs. loser mcts_h_ege0.1 (194)

round 4
winner mcts_h_ege1.0 (229) vs. loser mcts_h_ege0.9 (177)

Winner: mcts_h_ege1.0
\end{verbatim}

\subsubsection{Kalah Tournament Winner Comparisons}




Table~\ref{tbl:kalah-toptourney} shows the results of the top Kalah playout winners.
Each data point in the following table includes 2000 games. As mentioned in the main part of the paper, draws due to unbalanced 
boards are not counted. 

\begin{table}[h!]
\begin{center}
\begin{tabular}{|c|c|cc|}
\hline
Player A & Player B                  & A Wins (\%)  & B Wins (\%)   \\ 
\hline
MCTS(fet$3$)   & MCTS(fet$4$)        & 245 (48.80)   & 257 (51.20)    \\
MCTS(ege$0.9$) & MCTS(ege$1.0$)      & 389 (45.55)   & 465 (54.45)    \\
\hline
MCTS(ege$0.9$) & MCTS(fet$3$)        &  74 (7.05)   & 976 (92.95)    \\
MCTS(ege$0.9$) & MCTS(fet$4$)        &  66 (6.01)   & 1032 (93.99)    \\
MCTS(ege$1.0$) & MCTS(fet$3$)        &  93 (9.25)   & 912 (90.75)    \\
MCTS(ege$1.0$) & MCTS(fet$4$)        &  71 (6.96)   & 949 (93.04)    \\
\hline
\end{tabular}
\end{center}
\caption{Kalah playout comparisons. \label{tbl:kalah-toptourney}}
\end{table}

\vspace{10cm}

\subsection{Breakthrough Playout Enhancements (using efMS evaluator)}

\subsubsection{Fixed Early Terminations Tournament}

\begin{verbatim}
round 1
winner mcts_h_fet1000 (115) vs. loser mcts_h_fet0 (85)
winner mcts_h_fet100 (117) vs. loser mcts_h_fet1 (83)
winner mcts_h_fet50 (108) vs. loser mcts_h_fet2 (92)
winner mcts_h_fet30 (138) vs. loser mcts_h_fet3 (62)
winner mcts_h_fet20 (129) vs. loser mcts_h_fet4 (71)
winner mcts_h_fet10 (129) vs. loser mcts_h_fet5 (71)
mcts_h_fet8 gets a by

round 2
winner mcts_h_fet8 (108) vs. loser mcts_h_fet1000 (92)
winner mcts_h_fet10 (112) vs. loser mcts_h_fet100 (88)
winner mcts_h_fet20 (128) vs. loser mcts_h_fet50 (72)
mcts_h_fet30 gets a by

round 3
winner mcts_h_fet30 (113) vs. loser mcts_h_fet8 (87)
winner mcts_h_fet20 (104) vs. loser mcts_h_fet10 (96)

round 4
winner mcts_h_fet20 (104) vs. loser mcts_h_fet30 (96)

Winner: mcts_h_fet20
\end{verbatim}


\subsubsection{Epsilon-greedy Playout Tournament}

\begin{verbatim}
round 1
winner mcts_h_ege0.0 (156) vs. loser mcts_h_ege1.0 (44)
winner mcts_h_ege0.05 (155) vs. loser mcts_h_ege0.9 (45)
winner mcts_h_ege0.1 (156) vs. loser mcts_h_ege0.8 (44)
winner mcts_h_ege0.15 (153) vs. loser mcts_h_ege0.7 (47)
winner mcts_h_ege0.2 (151) vs. loser mcts_h_ege0.6 (49)
winner mcts_h_ege0.3 (119) vs. loser mcts_h_ege0.5 (81)
mcts_h_ege0.4 gets a by

round 2
winner mcts_h_ege0.0 (115) vs. loser mcts_h_ege0.4 (85)
winner mcts_h_ege0.05 (119) vs. loser mcts_h_ege0.3 (81)
winner mcts_h_ege0.1 (125) vs. loser mcts_h_ege0.2 (75)
mcts_h_ege0.15 gets a by

round 3
winner mcts_h_ege0.15 (103) vs. loser mcts_h_ege0.0 (97)
winner mcts_h_ege0.1 (110) vs. loser mcts_h_ege0.05 (90)

round 4
winner mcts_h_ege0.1 (108) vs. loser mcts_h_ege0.15 (92)

Winner: mcts_h_ege0.1
\end{verbatim}

\subsubsection{Dynamic Early Terminations Tournament}

\begin{verbatim}
round 1
winner mcts_h_det1.0 (121) vs. loser mcts_h_det0.1 (79)
winner mcts_h_det0.95 (115) vs. loser mcts_h_det0.15 (85)
winner mcts_h_det0.9 (120) vs. loser mcts_h_det0.2 (80)
winner mcts_h_det0.25 (101) vs. loser mcts_h_det0.85 (99)
winner mcts_h_det0.3 (119) vs. loser mcts_h_det0.8 (81)
winner mcts_h_det0.35 (117) vs. loser mcts_h_det0.75 (83)
winner mcts_h_det0.4 (107) vs. loser mcts_h_det0.7 (93)
winner mcts_h_det0.45 (132) vs. loser mcts_h_det0.65 (68)
winner mcts_h_det0.5 (106) vs. loser mcts_h_det0.6 (94)
mcts_h_det0.55 gets a by

round 2
winner mcts_h_det0.55 (129) vs. loser mcts_h_det1.0 (71)
winner mcts_h_det0.5 (124) vs. loser mcts_h_det0.95 (76)
winner mcts_h_det0.45 (115) vs. loser mcts_h_det0.9 (85)
winner mcts_h_det0.25 (134) vs. loser mcts_h_det0.4 (66)
winner mcts_h_det0.3 (108) vs. loser mcts_h_det0.35 (92)

round 3
winner mcts_h_det0.3 (133) vs. loser mcts_h_det0.55 (67)
winner mcts_h_det0.25 (120) vs. loser mcts_h_det0.5 (80)
mcts_h_det0.45 gets a by

round 4
winner mcts_h_det0.3 (135) vs. loser mcts_h_det0.45 (65)
mcts_h_det0.25 gets a by

round 5
winner mcts_h_det0.3 (106) vs. loser mcts_h_det0.25 (94)

Winner: mcts_h_det0.3
\end{verbatim}

\subsubsection{Breakthrough Tournament Winner Comparisons (using efMS)}



Table~\ref{tbl:bt-efMS-toptourney} shows the results of the top Breakthrough playout winners using efMS.

\begin{table}[h!]
\begin{center}
\begin{tabular}{|c|c|ccc|}
\hline
Player A & Player B                             & A Wins (\%)  & B Wins (\%)  & Ties \\ 
\hline
MCTS(ege$0.1$)              & MCTS(ipp,fet$20$)          & 557 (55.7)   & 443 (44.3)   & 0    \\
MCTS(ege$0.1$)              & MCTS(ipp,fet$4$)           & 768 (76.8)   & 232 (23.2)   & 0    \\
MCTS(ege$0.1$)              & MCTS(ipp,det$0.3$)         & 815 (81.5)   & 185 (18.5)   & 0    \\
MCTS(ipp,det$0.3$)          & MCTS(ipp,fet$20$)          & 719 (71.9)   & 281 (28.1)   & 0    \\
MCTS(ipp,det$0.3$)          & MCTS(ipp,fet$4$)           & 715 (71.5)   & 285 (28.5)   & 0    \\
\hline
MCTS(ege$0.1$,det$0.3$) & MCTS(ege$0.1$)         & 552 (55.2)   & 448 (44.8)   & 0    \\
MCTS(ege$0.1$,det$0.5$) & MCTS(ege$0.1$)         & 748 (74.8)   & 252 (25.2)   & 0    \\
MCTS(ege$0.1$,det$0.7$) & MCTS(ege$0.1$)         & 613 (61.3)   & 387 (38.7)   & 0    \\
MCTS(ege$0.1$,det$0.5$) & MCTS(ipp,fet$20$,det$0.5$) & 633 (63.3)   & 367 (36.7)   & 0    \\
\hline
\end{tabular}
\end{center}
\caption{Breakthrough playout comparisons using efMS. \label{tbl:bt-efMS-toptourney}}
\end{table}

\subsection{Breakthrough Playout Enhancements (using efLH evaluator)}

\subsubsection{Fixed Early Terminations Tournament}

\begin{verbatim}
round 1
winner mcts_h_efv1_fet0 (118) vs. loser mcts_h_efv1_fet1000 (82)
winner mcts_h_efv1_fet1 (129) vs. loser mcts_h_efv1_fet100 (71)
winner mcts_h_efv1_fet2 (113) vs. loser mcts_h_efv1_fet50 (87)
winner mcts_h_efv1_fet3 (101) vs. loser mcts_h_efv1_fet30 (99)
winner mcts_h_efv1_fet20 (121) vs. loser mcts_h_efv1_fet4 (79)
winner mcts_h_efv1_fet5 (100) vs. loser mcts_h_efv1_fet16 (100)
winner mcts_h_efv1_fet8 (102) vs. loser mcts_h_efv1_fet12 (98)
mcts_h_efv1_fet10 gets a by

round 2
winner mcts_h_efv1_fet10 (108) vs. loser mcts_h_efv1_fet0 (92)
winner mcts_h_efv1_fet8 (112) vs. loser mcts_h_efv1_fet1 (88)
winner mcts_h_efv1_fet5 (115) vs. loser mcts_h_efv1_fet2 (85)
winner mcts_h_efv1_fet20 (123) vs. loser mcts_h_efv1_fet3 (77)

round 3
winner mcts_h_efv1_fet20 (110) vs. loser mcts_h_efv1_fet10 (90)
winner mcts_h_efv1_fet8 (101) vs. loser mcts_h_efv1_fet5 (99)

round 4
winner mcts_h_efv1_fet8 (106) vs. loser mcts_h_efv1_fet20 (94)

Winner: mcts_h_efv1_fet8
\end{verbatim}

\subsubsection{Epsilon-greedy Playout Tournament}

\begin{verbatim}
round 1
winner mcts_h_efv1_ege1.0 (136) vs. loser mcts_h_efv1_ege0.0 (64)
winner mcts_h_efv1_ege0.9 (121) vs. loser mcts_h_efv1_ege0.05 (79)
winner mcts_h_efv1_ege0.1 (110) vs. loser mcts_h_efv1_ege0.8 (90)
winner mcts_h_efv1_ege0.7 (103) vs. loser mcts_h_efv1_ege0.15 (97)
winner mcts_h_efv1_ege0.6 (104) vs. loser mcts_h_efv1_ege0.2 (96)
winner mcts_h_efv1_ege0.3 (100) vs. loser mcts_h_efv1_ege0.5 (100)
mcts_h_efv1_ege0.4 gets a by

round 2
winner mcts_h_efv1_ege1.0 (122) vs. loser mcts_h_efv1_ege0.4 (78)
winner mcts_h_efv1_ege0.3 (101) vs. loser mcts_h_efv1_ege0.9 (99)
winner mcts_h_efv1_ege0.6 (116) vs. loser mcts_h_efv1_ege0.1 (84)
mcts_h_efv1_ege0.7 gets a by

round 3
winner mcts_h_efv1_ege0.7 (102) vs. loser mcts_h_efv1_ege1.0 (98)
winner mcts_h_efv1_ege0.3 (105) vs. loser mcts_h_efv1_ege0.6 (95)

round 4
winner mcts_h_efv1_ege0.3 (110) vs. loser mcts_h_efv1_ege0.7 (90)

Winner: mcts_h_efv1_ege0.3
\end{verbatim}

\subsection{Breakthrough Tournament Winner Comparisons (using efLH)}


Table~\ref{tbl:bt-efLH-toptourney} shows the results of the top Breakthrough playout winners using efLH.

\begin{table}[h!]
\begin{center}
\begin{tabular}{|c|c|ccc|}
\hline
Player A & Player B                  & A Wins (\%)  & B Wins (\%)  & Ties \\ 
\hline
MCTS(ipp,fet$8$)   & MCTS(ipp,fet$20$)       & 514 (51.4)   & 486 (48.6)   & 0    \\
MCTS(ege$0.3$)     & MCTS(ipp,fet$0.7$)      & 510 (51.0)   & 490 (49.0)   & 0    \\
\hline
MCTS(ege$0.3$) & MCTS(ipp,fet$8$)        & 354 (35.4)   & 646 (64.6)   & 0    \\
MCTS(ege$0.3$) & MCTS(ipp,fet$20$)       & 340 (34.0)   & 660 (66.0)   & 0    \\
MCTS(ege$0.7$) & MCTS(ipp,fet$8$)        & 255 (25.5)   & 745 (74.5)   & 0    \\
MCTS(ege$0.7$) & MCTS(ipp,fet$20$)       & 194 (19.4)   & 806 (80.6)   & 0    \\
\hline
\end{tabular}
\end{center}
\caption{Breakthrough playout comparisons using efLH. \label{tbl:bt-efLH-toptourney}}
\end{table}

\end{document}